\documentclass{article}
\usepackage{spconf,amsmath,graphicx,hyperref}
\usepackage{cite}
\usepackage{amssymb,amsfonts}
\usepackage{algorithmic}
\usepackage{textcomp}
\usepackage{pgfplots}
\usepackage{xcolor}
\usepackage{enumitem}
\usepackage{booktabs}
\usepackage{CJKutf8}
\usepackage{makecell}
\usepackage{multirow}
\usepackage{multicol}

\usepackage{graphicx}
\newcommand{\heart}{\ensuremath\heartsuit}

\title{CantoASR: Prosody-Aware ASR-LALM Collaboration for Low-Resource Cantonese}
\name{Dazhong Chen$^{\star\dagger}$, Yi-Cheng Lin$^{\ddagger}$, Yuchen Huang$^{\dagger}$, Ziwei Gong$^{\heart}$, Di Jiang$^{\S}$, Zeying Xie$^{\S}$, Yi R. (May) Fung$^{\dagger}$}

\address{$^{\star}$ The Chinese University of Hong Kong, Shenzhen   $^{\dagger}$ Hong Kong University of Science and Technology \\
$^{\ddagger}$ National Taiwan University, Taiwan  ~~~ $^{\heart}$Columbia University~~~
  $^{\S}$ WeBank Co., Ltd., Shenzhen, China}
\begin{document}
\ninept
\maketitle
%



\begin{abstract}
Automatic speech recognition (ASR) is critical for language accessibility, yet low-resource Cantonese remains challenging due to limited annotated data, six lexical tones, tone sandhi, and accent variation. Existing ASR models, such as Whisper, often suffer from high word error rates. Large audio-language models (LALMs), in contrast, can leverage broader contextual reasoning but still require explicit tonal and prosodic acoustic cues. We introduce \textit{CantoASR}, a collaborative ASR–LALM error correction framework that integrates forced alignment for acoustic feature extraction, a LoRA-finetuned Whisper for improved tone discrimination, and an instruction-tuned Qwen-Audio for prosody-aware correction. Evaluations on spontaneous Cantonese data show substantial CER gains over Whisper-Large-V3. These findings suggest that integrating acoustic cues with LALM reasoning provides a scalable strategy for low-resource tonal and dialectal ASR.
\end{abstract}

\vspace{-8pt}
\section{Introduction}
\vspace{-8pt}
\label{sec:intro}
Automatic Speech Recognition (ASR) for low-resource tonal languages has long been a core challenge in speech signal processing, driven by the growing demand for inclusive speech technology in multilingual societies \cite{ngueajio2022hey, 10887615}. Among these languages, Cantonese is especially important: spoken by over 80 million people worldwide, it features six lexical tones, complex tone sandhi rules, and widespread code-switching with English in daily communication. Accurate ASR for Cantonese is crucial not only for advancing speech technology for Cantonese-speaking communities, but also for building frameworks that generalize to other under-resourced tonal languages such as Hokkien, Vietnamese, and many African languages.

Despite its importance, robust recognition of tonal languages remains difficult because lexical tone, encoded by pitch height, contour, and duration, is easily distorted by noise or rapid speech. Standard augmentations (e.g., noise injection, speed perturbation) improve robustness to acoustic variability but cannot produce linguistically valid tonal variants \cite{s25144288}. As a result, even strong baselines such as Whisper \cite{whisper} exhibit systematic tone confusions, especially in conversational Cantonese with checked-tone shortening \cite{5684487}. Building tone-aware supervision further requires costly manual annotation, leaving a gap between acoustic features and tone-level phonological rules.

Recent advances in Large Audio-Language Models (LALMs) offer a new pathway to address this gap \cite{survey_sllm, yang2024buildingtaiwanesemandarinspoken}. Unlike conventional ASR systems, LALMs jointly process speech and text, combining strong acoustic understanding with the reasoning capability of large language models \cite{audio_language_modeling, desta2.5_audio, dynamic_superb}. This dual competence makes them particularly suitable for tonal languages, where resolving ambiguities requires both fine-grained prosodic analysis and phonological reasoning. Leveraging LALMs, therefore, provides a promising strategy for bridging the gap between raw acoustic cues and high-level linguistic correction, opening a scalable solution for Cantonese ASR and related tonal languages.

Hence, we propose CantoASR, a collaborative ASR–LALM framework that integrates acoustic-prosodic cues with phonological reasoning. Our method proceeds in four stages. First, we preprocess training corpora with noise augmentation and forced alignment, enabling the extraction of tone-relevant acoustic features such as F0, slope, and duration without manual labels. Second, we fine-tune Whisper-Large-V3 \cite{whisper} with parameter-efficient LoRA \cite{lora} adaptation, improving recognition of Cantonese while preserving efficiency. Third, we construct an instruction dataset at the word level, pairing each token with explicit correction prompts derived from its acoustic descriptors. This dataset is used to fine-tune Qwen2-Audio \cite{qwen2} for phonology-aware error correction, targeting common tone confusions and accent-related variations. Finally, during inference, we filter out high-confidence words, apply staged correction (tone and accent), and verify corrections through both acoustic and semantic checks.

Our contributions are fourfold. First, we introduce a framework that directly links acoustic-prosodic measurements to phonological rules via instruction tuning. Second, we will release the first Cantonese ASR error correction instruction tuning dataset, providing a valuable resource for the community. Third, we demonstrate how confidence filtering and multi-stage reasoning reduce overcorrection while improving tonal accuracy. Finally, experiments on Cantonese corpora show significant improvements over Whisper in both word error rate and tone recognition, highlighting a scalable recipe for other low-resource tonal languages such as Hokkien and Vietnamese.

\vspace{-10pt}
\section{Related Work}
\vspace{-8pt}
\begin{figure*}
    \centering
    \includegraphics[width=1.0\linewidth]{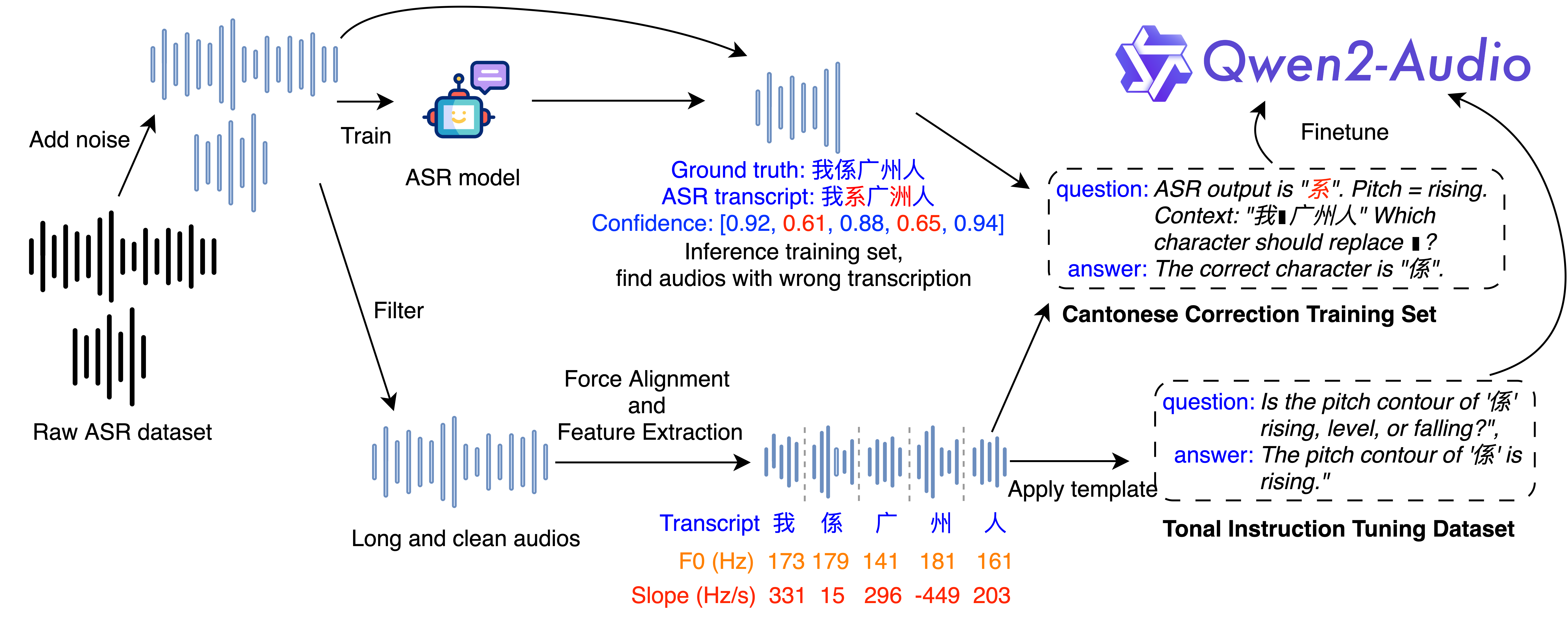} 
    \vspace{-18pt}
    \caption{Overview of CantoASR. The pipeline integrates prosody cues (F0, slope, duration) into LALM ASR error correction to build a tonal instruction-tuning dataset, and leverages ASR error patterns to build a Cantonese correction dataset.}
    \label{fig:workflow}
    \vspace{-10pt}
\end{figure*}

ASR error correction for tonal languages has attracted increasing attention, as conventional end-to-end systems often fail to capture tonal distinctions essential for meaning. Early neural approaches framed correction as a sequence-to-sequence “spelling correction” problem, using Transformer models to map noisy ASR hypotheses to corrected text \cite{wang20p_interspeech, zhao21_interspeech, ma23e_interspeech}. While effective for character-level substitutions, these methods typically ignored tone, producing outputs that were linguistically fluent but phonetically inconsistent with the speech signal \cite{fang-etal-2022-non}.

To mitigate tonal confusions, researchers have integrated phonological features into correction models \cite{fan23b_interspeech, dong2024pronunciation}. For Mandarin, PhVEC \cite{fang-etal-2022-non} appends pinyin tokens to input, enabling flexible edits while preserving pronunciation. Pinyin Regularization \cite{tang24_interspeech} adds tone-marked pinyin to hypotheses to reduce homophone errors missed by text-only models. Yet these methods remain Mandarin-focused and often simplify tonal complexity, while Cantonese presents a greater challenge with six tones and accent-driven variation.

LLMs have shifted ASR post-processing toward instruction-following and reasoning \cite{10389673}. RLLM-CF\cite{rllm_cf} proposed a three-stage pipeline that detects low-confidence words, applies iterative correction, and verifies outputs, reducing the overcorrection common in free-form rewriting. Ma et al. \cite{10930744} explored ASR error correction with LLMs using both 1-best and N-best hypotheses, showing gains over text-only baselines. Tang et al. \cite{10890161} introduced full-text correction for Chinese speech recognition, highlighting the benefits of long-context finetuning. While effective, these works rely solely on textual inputs and do not incorporate acoustic cues such as F0 contours or duration, limiting their ability to handle tonal ambiguities grounded in the signal.

\vspace{-10pt}

\section{Method}

Figure~\ref{fig:workflow} presents an overview of the CantoASR workflow, which integrates ASR finetuning, phonology-aware instruction tuning, and inference-time correction. The pipeline begins with preprocessing and feature extraction, proceeds through the construction of the Tonal Instruction Tuning Dataset, and culminates in a collaborative inference stage where Whisper provides candidate transcriptions and Qwen2-Audio-7B-Instruct applies targeted corrections. Each component is designed to explicitly incorporate tonal and prosodic information, ensuring that acoustic cues are preserved and effectively mapped to phonological rules.

\vspace{-8pt}
\subsection{Data preprocessing}
\label{subsec:intro}
We use the Common Voice Cantonese \cite{common_voice}, MCE \cite{mce}, and MDCC \cite{mdcc} datasets for ASR training, development, and testing. The data is split into 80\% for training, 10\% for validation, and 10\% for testing. 
All datasets are augmented with additive noise at 20 dB SNR to improve robustness. 
Specifically, for \textbf{70\%} of the audio data, we inject background speech from the MUSAN corpus \cite{musan}, where \emph{non-Cantonese conversational speech} is selected to simulate the classic ``cocktail party'' multi-speaker scenario \cite{cherry1953cocktail}. 
For the remaining \textbf{30\%}, we inject noise from the DEMAND dataset \cite{thiemann2013demand}, which contains recordings from 18 distinct real-world acoustic environments. 
We align the noise environments with utterances according to their dialogue topic labels; for example, weather-related discussions are more likely to occur in public places such as cafes or bus stations, and thus ambient sounds from these environments are injected into the corresponding audio to better simulate Cantonese daily conversational contexts.

\vspace{-8pt}
\subsection{ASR Finetuning}
\label{ssec:asr_finetune}
For acoustic modeling, we adopt Whisper-Large-V3~\cite{whisper} as the foundation model, motivated by its strong cross-lingual generalization and robustness to diverse acoustic conditions. To improve robustness while preserving tonal cues, we perform finetuning with LoRA adapters on the \emph{union} of the original clean training set and a lightly noised copy produced by our noise-injection pipeline (Sec.~\ref{subsec:intro}). The finetuned model outputs (i) raw transcriptions and (ii) word-level confidence by aggregating token posteriors over each word’s subword span, which are used for downstream correction.



\vspace{-8pt}
\subsection{Tonal Instruction Tuning Dataset Construction}
\label{subsec:tonal_instruction}
To provide phonology-aware supervision, we construct an instruction finetuning dataset from the processed ASR corpora. 
Audio segments shorter than five seconds or with SNR $< 10$\,dB are excluded. 
Forced alignments are obtained using the Montreal Forced Aligner (MFA)~\cite{mcauliffe17_interspeech}, with a Cantonese acoustic model trained on the \texttt{zh-HK} split of Common Voice~\cite{common_voice}. 
The pronunciation lexicon is derived from CharsiuG2P~\cite{zhu2022byt5modelmassivelymultilingual} and PyCantonese~\cite{lee-etal-2022-pycantonese}, covering 2450 daily-use Cantonese characters.

We perform alignment at the \emph{character/word} level and extract prosodic descriptors using Parselmouth~\cite{parselmouth}, including (i) speaker-normalized $F_0$ 
, (ii) $F_0$ slope within each token span, and (iii) token duration and pitch stability. 
These features are paired with base-tone labels (from Jyutping-to-tone mapping) to form supervision.

We extract continuous prosodic features (mean $F_0$, $F_0$ slope, and speaking rate) and discretize them into categories by splitting values at the 25\%, 25–75\%, and 75\% quantiles computed from the training set. This yields bins such as \emph{low/medium/high} ($F_0$), \emph{rising/flat/falling} ($F_0$ slope), and \emph{slow/moderate/fast} (rate). 

Each sample contains both \textbf{closed-form classification} and \textbf{open-form reasoning} prompts. 
Closed-form asks for explicit categories (e.g., “Is the pitch contour of  ‘\begin{CJK}{UTF8}{bsmi}係\end{CJK}’ rising, level, or falling?” → “Pitch~rising”), 
while open-form integrates multiple descriptors (e.g., “Analyze ‘\begin{CJK}{UTF8}{bsmi}係\end{CJK}’: Tone~6, Pitch~rising, Slope~flat, Speed~fast. Explain how these reflect Cantonese phonology.”).  
For open-form answers, we construct explanatory text by mapping the discrete bins back to phonological interpretations (e.g., “rising pitch with a flat slope corresponds to Tone~6 in Cantonese”), thereby providing reasoning-style supervision rather than a single categorical label.
\vspace{-12pt}

\subsection{Qwen2-Audio Finetuning for Cantonese Correction}
\label{subsec:qwen2_audio_finetune}
To support phonology-aware correction, we construct an instruction finetuning dataset for every word in the ``Tonal Instruction Tuning Dataset,'' ensuring that each lexical item is paired with explicit correction instructions and audio segments with timestamps. This allows the model to learn systematic mappings between acoustic descriptors, phonological rules, and transcription refinements.

One major source of error involves tone confusion, which occurs when Whisper mislabels tones due to overlapping ranges of fundamental frequency ($F_0$) or similar slope trajectories. For example, an instruction may be phrased as: ``Based on the acoustic features ($F_0$ = 153.6 Hz, slope = 128.7 Hz/s), correct the tone error in the word ‘\begin{CJK}{UTF8}{gbsn}系\end{CJK}’.'' This explicit linking of acoustic cues to tonal categories teaches the model to resolve ambiguities and enforce the correct rising contour, thereby reducing systematic substitution errors across tonal categories. A second category of error stems from accent adaptation, where regional pronunciation patterns deviate from canonical tone realizations. An example instruction is: ``For the Hong Kong Cantonese accent (Tone 2 slope reduced by 13\%), correct the recognition result of the word ‘\begin{CJK}{UTF8}{gbsn}系\end{CJK}’.'' By including accent-aware instructions, the dataset enables the model to reinterpret slope reductions as legitimate accentual variation rather than as transcription error, thereby improving robustness across speaker populations.
\vspace{-12pt}

\subsection{Inference strategy}
\vspace{-2pt}
The inference stage of CantoASR is designed to improve transcription reliability through three steps: confidence-based filtering, staged correction, and layered consistency verification.

First, we analyze character-level confidence scores from Whisper. These scores are derived from the token-level posterior probabilities produced during decoding, aggregated over the subword units of each word. Tokens with confidence $\geq 0.7$ are preserved, while those below 0.7 are passed to the correction module. For example, in the transcription “\begin{CJK}{UTF8}{gbsn}我系广洲人\end{CJK}” with scores [0.92, 0.61, 0.88, 0.65, 0.94], only “\begin{CJK}{UTF8}{gbsn}系\end{CJK}” and “\begin{CJK}{UTF8}{gbsn}洲\end{CJK}” are targeted for correction, reducing redundant computation.

Secondly, the targeted words are sent for LLM correction in a linguistically motivated order: tone correction, then accent adaptation. This stepwise approach minimizes error propagation. In the example above, “\begin{CJK}{UTF8}{gbsn}系\end{CJK}” is first corrected to “\begin{CJK}{UTF8}{bsmi}係 \end{CJK}” (tone), followed by accent verification, with no further modifications required.

Finally, we apply two layers of validation. Acoustic checks confirm that corrected words match target tonal features (e.g., $F_0$ range and slope for Tone 6 in “\begin{CJK}{UTF8}{bsmi}係\end{CJK}”). Semantic checks, implemented with a lightweight Cantonese language model (CantoneseLLMChat-v1.0-7B \cite{cheng2025hkcantoevalbenchmarkevaluatingcantonese}), ensure contextual plausibility. This layer is particularly effective for homophone disambiguation and place-name correction, such as mapping “\begin{CJK}{UTF8}{gbsn}广洲\end{CJK}” to “\begin{CJK}{UTF8}{gbsn}广州\end{CJK}” (the correct spelling for “Guangzhou”, a major city in China).

\begin{table*}[t]
\centering
\caption{Cantonese ASR post-correction. 
\textbf{Text-only}: models see only ASR text. 
\textbf{Audio-conditioned}: models take audio+text. Performance is in CER (\%), lower is better.
\textbf{Boldface} indicates the best result in each column.}
\vspace{3pt}

\resizebox{\textwidth}{!}{%
\begin{tabular}{ll|ccc|ccc|ccc|cc}
\toprule
\multirow{2}{*}{\textbf{Category}} & \multirow{2}{*}{\textbf{Model}} 
& \multicolumn{3}{c|}{\textbf{Common Voice}} 
& \multicolumn{3}{c|}{\textbf{MCE}} 
& \multicolumn{3}{c|}{\textbf{MDCC}} 
& \multicolumn{2}{c}{\textbf{Overall}} \\
\cmidrule(lr){3-5}\cmidrule(lr){6-8}\cmidrule(lr){9-11}\cmidrule(lr){12-13}
&  & 1-best & 5-best & 10-best
   & 1-best & 5-best & 10-best
   & 1-best & 5-best & 10-best
   & Avg. CER &\\
\midrule
\multirow{3}{*}{\textbf{Text-only LLMs}} 
& GPT-4o & 31.96 & 30.36 & 28.76 & 32.23 & 30.61 & 29.00 & 25.37 & 20.44 & 18.24 & 27.3 \\
& LLaMA-3-8B & 57.95 & 55.05 & 52.15 & 58.25 & 55.34 & 52.43 & 62.8 & 57.39 & 52.91 & 55.8 \\
& Gemini-2.5-Pro & 63.78 & 60.59 & 57.40 & 63.68 & 60.50 & 57.32 & 61.32 & 56.04 & 51.26 & 59.0 \\
& gpt-oss & 27.95 & 26.55 & 25.15 & 28.14 & 26.73 & 25.32 & 50.7 & 45.08 & 40.37 & 32.8 \\

\midrule
\multirow{5}{*}{\textbf{LALM}} 
& SALMONN-7B & 40.20 & 38.19 & 36.18 & 40.09 & 38.08 & 36.08 & 38.09 & 33.2 & 30.11 & 36.7 \\
& Qwen2-Audio-7B & 16.64 & 15.80 & 14.97 & 16.54 & 15.71 & 14.89 & 14.53 & 11.54 & 9.51 & 14.5 \\
& DeSTA2.5-Audio & 29.67 & 28.18 & 26.70 & 30.19 & 28.68 & 27.17 & 33.58 & 28.4 & 25.96 & 28.5 \\
& Phi-4-Multimodal & 29.11 & 27.65 & 26.20 & 29.13 & 27.68 & 26.22 & 42.83 & 37.52 & 32.08 & 30.8 \\
\midrule
\textbf{Ours} 
& \textbf{CantoASR} 
& \textbf{12.13} & \textbf{10.81} & \textbf{10.52} 
& \textbf{12.16} & \textbf{10.53} & \textbf{10.25} 
& \textbf{12.12} & \textbf{11.27} & \textbf{10.98} 
& \textbf{11.19}\\
\bottomrule
\end{tabular}}
\label{tab:baseline_results}
\vspace{-14pt}
\end{table*}

\vspace{-10pt}

\section{Experiment Setup}
\noindent\textbf{Details on Whisper finetuning.}
We apply parameter-efficient adaptation to Whisper using \emph{AdaLoRA}, inserting low-rank adapters into the attention projection matrices (q, k, v, o). The adapters start with rank $r=12$ and are gradually reduced to a target rank of $r=4$, with a scaling factor $\alpha=32$ and dropout rate of 0.1. This setup keeps the number of trainable parameters very small, at only about 0.5--0.8\% of the full model. We optimize with AdamW ($10^{-4}$ learning rate, cosine schedule, 200 warmup steps, weight decay 0.01, label smoothing 0.1, gradient clipping 1.0), using mixed precision and an effective batch size of 16.



\noindent\textbf{Details on Qwen2-Audio finetuning.} We finetune \emph{Qwen2-Audio-7B-Instruct} using 4-bit NF4 quantization (BitsAndBytes) together with LoRA adapters placed on both attention and MLP projection layers. The adapters have rank $r=16$, scaling factor $\alpha=32$, dropout 0.05, and no bias, which keeps the number of trainable parameters low ($\approx$0.6–0.8\%). Training is carried out with 8-bit paged AdamW ($10^{-4}$ learning rate, 200 warmup steps, gradient clipping at 1.0), FP16 mixed precision, and an effective batch size of 8 (per-device batch size 2 with gradient accumulation of 4). We train for 3 epochs.
\vspace{-12pt}
\section{Experiments and Results}
\label{sec:exp} 
\vspace{-6pt}
\subsection{Datasets and Metrics}
We evaluate on three Cantonese benchmarks: (1) \textbf{Common Voice Cantonese (CV)}~\cite{common_voice}, an open-source read-speech corpus; (2) \textbf{MCE}~\cite{mce}, a low-resource collection (\raisebox{.5pt}{\texttildelow}10k utterances) with conversational and read speech; and (3) \textbf{Multi-Domain Cantonese Corpus (MDCC)}~\cite{mdcc}, a multi-domain set covering spontaneous speech and accent variation. We report \textbf{Character Error Rate (CER)} in \%, and average across test sets.
\vspace{-4pt}
\subsection{Baseline Comparison}\label{sec:baseline}
For each test utterance, the Cantonese finetuned Whisper in Sec.~\ref{ssec:asr_finetune} produces an \emph{$N$-best} list via standard beam search (\,$N\!\in\!\{1,5,10\}$\,). We then treat each baseline \emph{text-only LLM} and \emph{LALM} as an \textbf{ASR error-correction module} that is prompted with the candidate transcripts: the model receives the $N$ hypotheses (and, for LALMs, the original audio) and is instructed to produce a single corrected transcript.

Table~\ref{tab:baseline_results} reports per-dataset CER and an overall macro average across datasets and $N$. Among \emph{text-only} correctors, ChatGPT-4o attains an overall CER of 27.3, whereas LLaMA-3-8B and Gemini-2.5-Pro remain markedly higher (overall 55.8 and 59.0), indicating limited robustness to Cantonese tonal ambiguity. Moving to \emph{audio-conditioned} models, Qwen2-Audio-7B substantially lowers errors (overall \textbf{14.5}), outperforming SALMONN-7B (36.7), DeSTA2.5-Audio (28.5), and Phi-4-Multimodal (30.8). 

Our \textbf{CantoASR} achieves the best overall CER of \textbf{11.19}, improving over the strongest baseline Qwen2-Audio-7B by $\sim$3.3 absolute points. Notably, CantoASR exhibits consistent gains as hypothesis diversity increases: CER decreases from \textbf{12.13} (1-best) to \textbf{10.81}/\textbf{10.52} (5/10-best) on Common Voice, with the same monotonic trend on MCE and MDCC. This pattern suggests that (i) exposing the corrector to richer candidate sets benefits tone-sensitive correction, and (ii) our phonology-aware design exploits complementary acoustic evidence that is not captured by text-only approaches.

\vspace{-4pt}
\subsection{Ablation Study}
To isolate the contribution of each component, we evaluate variants in Table~\ref{tab:ablation_results}:
\begin{table}[tbp]
\centering
\caption{Ablation study on Cantonese ASR (CER, \%). ``w/o'' indicates removing finetuning at ASR or LLM stage.}
\label{tab:ablation_results}
\resizebox{0.5\textwidth}{!}{%
\begin{tabular}{lcccc}
\toprule
\textbf{Model Variant} & \textbf{CV} & \textbf{MCE} & \textbf{MDCC} & \textbf{Avg. CER} \\
\midrule
ASR only (w/o finetune)             & 21.21 & 19.82 & 22.53 & 21.24 \\
ASR-Finetune (w/o LLM)              & 15.22 & 14.73 & 16.84 & 15.64 \\
ASR-Finetune + LLM (w/o LLM FT)     & 16.83 & 15.74 & 17.95 & 16.86 \\
ASR-Finetune + LLM-Finetune         & 11.22 & 10.83 & 12.64 & 11.55 \\
\;\;\;\;+ CLLM (semantic correction) & \textbf{10.81} & \textbf{10.52} & \textbf{12.13} & \textbf{11.19} \\
\bottomrule
\end{tabular}}
\vspace{-15pt}
\end{table}
\begin{itemize}[topsep=2pt, itemsep=1pt, parsep=0pt, partopsep=0pt,
                leftmargin=1.5em, labelsep=0.6em]
    \item \textbf{ASR only (w/o finetune)}: Whisper-Large baseline.
    \item \textbf{ASR-Finetune (w/o LLM)}: LoRA finetuning on Whisper improves robustness under noisy conditions, reducing Avg.\ CER to $15.64\%$ ($-26.4\%$ rel.).
    \item \textbf{ASR-Finetune + LLM (w/o LLM FT)}: Adding a non-finetuned LLM for error correction yields $16.86\%$ Avg.\ CER, showing some phonology-aware benefit but limited without instruction tuning.
    \item \textbf{ASR-Finetune + LLM-Finetune}: With instruction tuning, Avg.\ CER drops to $11.55\%$, confirming the role of explicit tonal supervision.
    \item \textbf{ASR-Finetune + LLM-Finetune + CLLM (Ours)}: Adding a lightweight Cantonese LLM for semantic validation achieves the best $11.19\%$ Avg.\ CER, effectively correcting homophones and place-name errors.
\end{itemize}
Thus, \textbf{ASR-Finetune} improves \emph{robustness}, \textbf{LLM-Finetune} strengthens \emph{acoustic/tonal correction}, and \textbf{Cantonese LLM} enhances \emph{semantic correction}. Their integration yields the strongest system.
\begin{table}[tbp]
\centering
\caption{Impact of constrained decoding on Qwen2-Audio-7B-Instruct-Finetune (CER, \%).}
\label{tab:constrained}
\resizebox{0.48\textwidth}{!}{%
\begin{tabular}{clcccccc}
\toprule
\textbf{Input} & \textbf{Decoding} & $\lambda$ & \textbf{CV} & \textbf{MCE} & \textbf{MDCC} & \textbf{Avg.} \\
\midrule
\multirow{3}{*}{1-best} 
 & Unconstrained & --    & 15.01 & 14.02 & 14.53 & 14.52 \\
 & N-best        & 0.50  & 14.02 & 13.01 & 13.54 & 13.52 \\
 & Lattice       & 0.55  & 13.53 & 12.52 & 13.01 & 13.02 \\
\midrule
\multirow{3}{*}{5-best} 
 & Unconstrained & --    & 12.02 & 11.01 & 11.54 & 11.52 \\
 & N-best        & 0.45  & 11.03 & 10.02 & 10.51 & 10.52 \\
 & Lattice       & 0.50  & 10.54 &  9.53 & 10.02 & 10.03 \\
\midrule
\multirow{3}{*}{10-best} 
 & Unconstrained & --    & 10.03 &  9.02 &  9.51 &  9.52 \\
 & N-best        & 0.40  &  9.04 &  8.03 &  8.52 &  8.53 \\
 & Lattice       & 0.45  &  8.55 &  7.54 &  8.03 &  8.04 \\
\bottomrule
\end{tabular}}
\vspace{-10pt}
\end{table}


\vspace{-12pt}
\subsection{Impact of Constrained Decoding}
\label{sec:constrained}
We evaluate three strategies for the correction model:
(1) \textbf{Unconstrained}: free-form decoding of the corrected text; 
(2) \textbf{$N$-best constrained}: correct each ASR hypothesis in the $N$-best list, then select the output by interpolation scoring
\begin{equation}
\label{eq:top1ofn}
s_i \;=\; \log p_\theta(\hat{y}_i \mid \text{inputs}) \;+\; \lambda \,\log p_{\mathrm{ASR}}(x_i),
\end{equation}
where $x_i$ is the $i$th ASR hypothesis, $\hat{y}_i$ its corrected form, and $\lambda$ is tuned on the dev set; 
(3) \textbf{Lattice constrained}: perform prefix-constrained generation on the ASR lattice, using the same interpolation form in Eq.~\eqref{eq:top1ofn} with the lattice path prior.
Table~\ref{tab:constrained} reports the best dev-tuned $\lambda$ per setting, and macro-averages the CER over the three datasets.

Compared with unconstrained decoding, $N$-best constraints reduce Avg.\ CER by \textbf{6--10\%} relative, while lattice constraints yield up to \textbf{15.5\%} relative gains (Table~\ref{tab:constrained}). 
These gains are complementary to the model-side improvements of CantoASR and are adopted as our default decoding setting for best downstream accuracy.
\vspace{-8pt}
\subsection{Discussion}
The results highlight three points:  
(1) Generic LLMs struggle with Cantonese tonal ambiguities, while audio-aware models adapt better.  
(2) ASR and LLM finetuning provide complementary gains: ASR improves robustness to noisy speech, and LLM tuning reduces tone confusions.  
(3) The Cantonese LLM validator contributes further improvements in semantic plausibility, especially for homophones and place names.  
These gains are complementary to constrained decoding (Sec.~\ref{sec:constrained}), which further reduces CER by 6--15\% relative.

\vspace{-6pt}
\section{Conclusion}
We presented CantoASR, an ASR and LALM collaboration that grounds error correction in acoustic prosody and phonological reasoning. The system combines LoRA finetuning of Whisper on clean and lightly noised data, a tonal instruction tuning dataset that links token-level descriptors to tone categories, and a Qwen2-Audio corrector with constrained decoding and a lightweight Cantonese validator. On three Cantonese benchmarks, CantoASR attains an overall CER of 11.19\%, improving over the strongest baseline Qwen2-Audio-7B at 14.5\% by about 23\% relative, and constrained decoding provides an additional 6 to 15.5\% relative reduction. The method requires no manual tone labels, scales to noisy conversational speech, and offers a practical template for other tonal languages. We will release the Cantonese ASR error correction instruction dataset to facilitate reproducibility and further research. Future work includes joint end-to-end training of ASR and correction, stronger modeling of code switching and accent variation, streaming and low-latency inference, and extending the framework to Hokkien and Vietnamese, given their similar tonal systems and prosodic features to Cantonese. This extension will help apply the framework to a broader range of tonal languages and dialects.


\section*{Acknowledgment}
This research was supported in part by WeBank (Grant WEB24EG01-L).

\bibliographystyle{IEEEbib}
\bibliography{strings,refs}
\end{document}